\documentclass{article}

\usepackage{adafedfr_arxiv}

\usepackage[utf8]{inputenc} 
\usepackage[T1]{fontenc}    
\usepackage{hyperref}       
\usepackage{url}            
\usepackage{booktabs}       
\usepackage{amsfonts}       
\usepackage{nicefrac}       
\usepackage{microtype}      
\usepackage{xcolor}         
\usepackage{algorithm}
\usepackage{epsfig}
\usepackage{amsmath}
\usepackage{amssymb}
\usepackage{booktabs}
\usepackage{bm}
\usepackage{multirow}
\usepackage{algorithmicx}
\usepackage{algpseudocode}
\usepackage{makecell}
\usepackage{textcomp}
\usepackage{bbding}
\usepackage{microtype}
\usepackage{array}
\usepackage{setspace}
\usepackage{threeparttable}
\usepackage{diagbox}
\usepackage{wrapfig}
\usepackage{float}
\usepackage{bbm}
\usepackage{url}
\usepackage{mathdots}

\title{
AdaFedFR: Federated Face Recognition with Adaptive Inter-Class Representation Learning}

\author{%
  David S.~Hippocampus\thanks{Use footnote for providing further information
    about author (webpage, alternative address)---\emph{not} for acknowledging
    funding agencies.} \\
  Department of Computer Science\\
  Cranberry-Lemon University\\
  Pittsburgh, PA 15213 \\
  \texttt{hippo@cs.cranberry-lemon.edu} \\
}

\begin{document}

\maketitle

\begin{abstract}
  With the growing attention on data privacy and communication security in face recognition applications, federated learning has been introduced to learn a face recognition model with decentralized datasets in a privacy-preserving manner. However, existing works still face challenges such as unsatisfying performance and additional communication costs, limiting their applicability in real-world scenarios. In this paper, we propose a simple yet effective federated face recognition framework called AdaFedFR, by devising an adaptive inter-class representation learning algorithm to enhance the generalization of the generic face model and the efficiency of federated training under strict privacy-preservation. In particular, our work delicately utilizes feature representations of public identities as learnable negative knowledge to optimize the local objective within the feature space, which further encourages the local model to learn powerful representations and optimize personalized models for clients. Experimental results demonstrate that our method outperforms previous approaches on several prevalent face recognition benchmarks within less than 3 communication rounds, which shows communication-friendly and great efficiency.
\end{abstract}

\section{Introduction}

\par Face recognition technology has been widely applied in personal identity verification systems. The significant advancements in this field can be attributed to the development of convolutional neural networks (CNNs) and the availability of large-scale training datasets ~\cite{guo2016ms, nech2017level, an2021partial, zhu2021webface260m}. To further improve the performance of face recognition systems, several advanced loss functions \cite{{sun2014deep}, {taigman2014deepface}, {schroff2015facenet}, {masi2018deep}, {deng2017marginal}, {liu2017sphereface}, {wang2018cosface}, {wang2018additive}, {deng2019arcface}, {deng2020sub}, {deng2021variational}, {huang2020curricularface}} have been proposed.
However, face recognition in real-world scenarios remains a significant challenge because of the large variability exhibited by face images under unconstrained conditions. Factors such as illumination variations, complex backgrounds, and image blurriness\cite{terhorst2020ser} contribute to this variability and make accurate face recognition more difficult. Moreover, the limited availability of industrial data and the domain gap between public and industrial datasets pose additional challenges. Models trained on publicly available datasets often fail to achieve satisfactory performance when deployed in real-world industrial applications.


\par To address above problem, an intuitive approach is to combine the training dataset with available decentralized data on local devices. However, due to the increasing privacy concerns and data protection regulations \cite{voigt2017eu}, face data with personal identification is restricted to be collected by the industry. Considering privacy issues, federated learning is applied to enable multiple users and parties to jointly learn a face recognition model without sharing the local data\cite{mcmahan2017communication}.  A general federated learning method, FedAvg\cite{mcmahan2017communication}, transfers the local updated models to the server, and aggregates the the local models to the updated global model. However, a key challenge in federated learning is the heterogeneity of data distribution, especially non-identically distributed (non-IID) data in real-world applications.
\par Recently, some significant studies\cite{{li2020federated}, {karimireddy2020scaffold}, {li2021model}}on image classification tasks have been proposed to address the non-IID issues. In face recognition, the non-IID data distribution can be extremely pronounced. Each local client typically has fewer identities compared to the dataset available on the server, resulting in a significant imbalance in data distribution. 
To the best of our knowledge, FedFace\cite{aggarwal2021fedface} is the first federated face recognition framework to explore improving the global aggregation process by uploading local class embeddings from clients to the server. Moreover, compared with FedFace, FedGC\cite{niu2022federated} applies gradient correction for federated face recognition with private embedding and enhances the privacy security by achieving differential privacy. Subsequently, Liu {\em et al.} proposes the FedFR\cite{liu2022fedfr} focuses on local training to avoid the transmission of local class embeddings, which utilizes large-scale publicly available data to regularize training on the local clients, resulting in performance improvement. However, this approach incurs additional computation costs, which limit its practicality and efficiency in real-world face recognition scenarios. 
\par In our work, we propose an effective and practical federated face recognition framework called AdaFedFR, 
aiming to performance enhancement and resource-efficient while ensuring privacy. Different from previous works, we utilize shareable mean feature embeddings of public identities as global class representations, rather than shared data, to 
improve the discriminative capability of generic face representations and the efficiency of the overall framework, simultaneously. Particularly, we incorporate a novel learnable $k$ negative contrastive objective that negatives from global representations are adaptively chosen based on their proximity to the positive samples in the embedding space (i.e., negative pairs), and maximizes the agreement of representation learned by the local model and the representation learned by the global model (i.e., positive pair) to mitigate client drift and ensure consistency with the global model.
Additionally, our framework simultaneously optimizes the user experience on local clients by adding an adapter behind local feature extractor. Furthermore, the differential privacy endorsement for transmitting parameters can enhance the security of our federated framework.
Empirical results demonstrate the effectiveness of AdaFedFR in terms of speeding up the convergence of the generic model and optimizing personalized models compared to previous approaches. Particularly, due to the reduced number of communication rounds required in federated learning, our method lowers the risk of privacy leakage.
\par To evaluate the generic performance of our proposed framework, extensive experiments are conducted on multiple face recognition benchmark datasets, including IJB-B\cite{whitelam2017iarpa} and IJB-C\cite{maze2018iarpa}. It demonstrates that AdaFedFR is superior to previous methods\cite{{liu2022fedfr}, {aggarwal2021fedface}}, and validates the effectiveness of the proposed AdaFedFR to alleviate the additional transmission and computation resource under the federated learning scenario. Our main contributions are summarized as follows:
\vspace{-0.2cm}
\begin{itemize}
    \item We propose a simple yet effective federated face recognition framework, AdaFedFR, significantly improving the applicability and efficiency of federated learning framework for face recognition tasks under strict privacy restrictions.
    \item We design a novel local objective that utilizes adaptive inter-class representation learning with publicly available class representations to improve the local updates and avoid overfitting to limited classes at each client.
    \item Extensive experiments on the different generic benchmarks demonstrate that our framework can achieve outperforming performance with negligible computation cost in comparison with previous methods. 
    \item Our method can achieve the best accuracy in less than three communication rounds, which enables AdaFedFR more friendly to be employed in real-world scenarios.
\end{itemize}

\section{Related Work}

\subsection{Face Recognition}
\indent Deep face recognition \cite{masi2018deep} has achieved satisfying performances and widely applied on a variety of situations, such as security and authentication, because of the large-scale datasets \cite{{cao2018vggface2}, {guo2016ms}, {zhu2021webface260m}}and widely usage of margin-based softmax loss functions \cite{{liu2016large}, {liu2017sphereface}, {wang2018additive}, {wang2018cosface}, {deng2019arcface}}. Despite previous well-designed loss functions in face recognition, such as SphereFace\cite{liu2017sphereface}, CosFace\cite{wang2018cosface}, and ArcFace\cite{deng2019arcface}, 
have enhanced the discriminative capability of face feature representation by adopting different forms of margin penalty, these methods do not directly apply for the federated learning framework since their compelling performance mainly results from centralized data. 
\subsection{Federated Learning}
\indent Federated Learning is a machine learning setting, which is proposed to learn a model with decentralized clients in a collaborative manner under data privacy and security \cite{li2021survey}. FedAvg\cite{mcmahan2017communication}simply optimizes the local training objective function and aggregates with a weighted average approach. Recent works 
on improving FedAvg can be summarized following two focuses: models aggregation at the server and local training optimization. FedProx\cite{li2020federated} and SCAFFOLD\cite{karimireddy2020scaffold} focus on local training, by introducing a proximal term to constraint the local model update and correcting the local updates with control variates. MOON\cite{li2021model} conducts contrastive learning in model-level. FedAwS\cite{yu2020federated} imposes a geometric regularization at the server to encourage class embeddings spreadout. However, these approaches are primarily applicable to small-scale datasets and suffer from privacy leakage with respect to face feature embeddings. 
\subsection{Federated Learning in Face Recognition}
\indent Recently, some works\cite{{aggarwal2021fedface}, {meng2022improving}, {niu2022federated}} on face recognition tasks have emerged to address the challenges of face model optimization in the federated learning setup. Fedface\cite{aggarwal2021fedface} is a relatively early work applying federated learning to face recognition, which introduces local class centers and a spreadout regularizer to separate the class embeddings from different clients. Meng {\em et al.} \cite{meng2022improving} enhances the privacy-preserving via communicating auxiliary information among clients with differential privacy. Niu {\em et al.} proposed a federated learning framework, FedGC\cite{niu2022federated}, exploring gradients correction via a softmax-based regularizer from a novel perspective of back propagation. To improve both generic and personalized face recognition models in a privacy-aware manner, FedFR\cite{liu2022fedfr} proposes a joint optimization framework by leveraging globally shareable data. However, the excessive communication and computation consumption still restricts their applications in various real-world scenes.
\par In this paper, we address the excessive communication and local computation consumption raised in recent method\cite{liu2022fedfr} and employ the inter-class representations learning based approach to enhance the federated training efficiency in real-world scenario. Besides, our proposed framework further improve the performance of generic and personalized face recognition model under higher privacy guarantee.

\section{Methodology}
This section provides our proposed framework for federated learning in face recognition system. Before illustrating the details of our method, we present the problem definition and the motivation first.
\subsection{Problem Setup}
\noindent \textbf{Face Recognition}
\indent For an open-set problem, face recognition model can be considered as a multi-class classifier which can separate different identities in the training set and learn discriminative representations in the feature space. Then, the similarity between feature embeddings represents whether two faces would belong to the identical class. Empirically, the practical face recognition model is trained on a large-scale dataset $D_S$ aiming at learning better feature embeddings. In our setting, the pre-trained generic face model is trained with the commonly used large margin cosine loss(LMCL) \cite{wang2018cosface}:
\begin{equation}\label{1}
 \mathcal{L}_{lmc} = \frac{1}{\left | D \right |}\sum_{i}{-\log{\frac{e^{s (\cos(\theta_{{y_i}, i}) - m)}}{e^{s (\cos(\theta_{{y_i}, i}) - m)} + \sum_{j \neq y_i}{e^{s \cos(\theta_{j, i})}}}}}, 
\end{equation}
subject to 
\begin{equation}\label{2}
\begin{split}
  w &= \frac{w^*}{\Vert{w^*}\Vert}, \\
  x &= \frac{x^*}{\Vert{x^*}\Vert}, \\
  cos(\theta_{j,i}) &= {w_j}^Tx_i,
\end{split}
\end{equation}

where $m$ is a fixed hyper-parameter introduced to control the magnitude of the cosine margin. $\left | D \right |$ is the number of training samples, $x_i$ is the $i$-th feature vector corresponding to the ground-truth class of $y_i$, the $w_j$ is the weight vector of the $j$-th class, and $\theta_j$ is the angle between $w_j$ and $x_i$.
\par In federated learning setup, dataset $D$ is always collected and stored on the client side, which cannot be shared with the constraint of data privacy.
\vspace*{0.3\baselineskip}  

\begin{figure*}
\begin{center}
\includegraphics[width=0.88\linewidth]{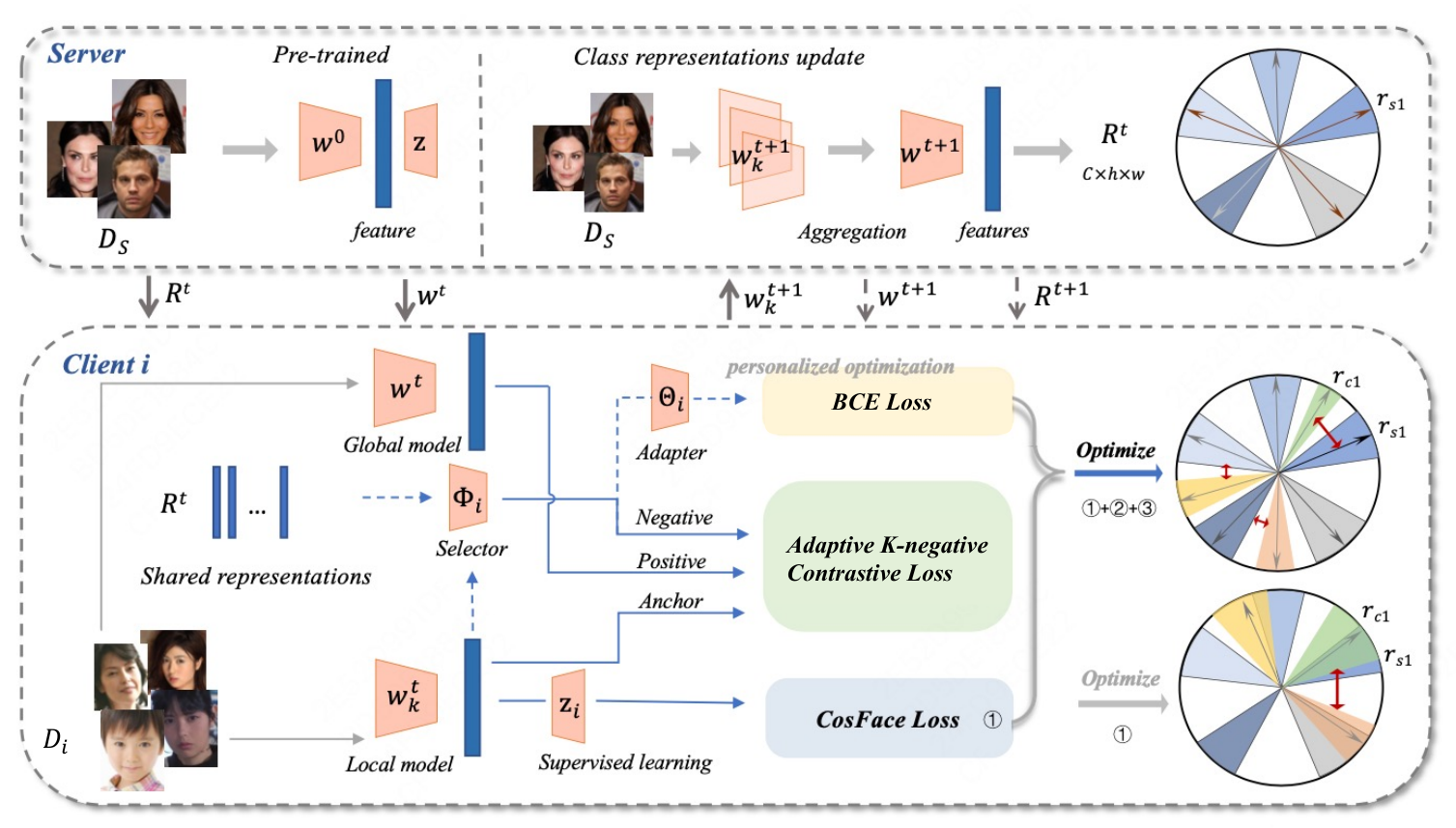}
\vspace{-0.2cm}
\end{center}
   \caption{Overview of our proposed federated face recognition framework. At each communication round, we optimize the local model with local data and global class representations by utilizing a combined objective consisting of cosface loss, adaptive k-negative contrastive loss and bce loss, which effectively enlarge the distances of different identities (i.e. $r_{c1}$ and $r_{s1}$) in feature space. }
\label{fig2}
\end{figure*} 

\par\noindent \textbf{Federated Learning}
\indent We consider a general federated learning setting consisting of one server and $N$ clients. With sufficient computing resources and a large scale of public face data, The server can always hold a well pre-trained face recognition model $w^0$, which can be sent to initialize the local model at clients. Let $D_i$ denote the local database at each client $C_i$, where $i\in\left \{ 1,2,...,N \right \}$.At the server, the goal is to learn a model over decentralized data at $N$ clients. A client $C_i$ participating in the federated process, aims to find an extractor $w^0$ to minimize a certain loss function. Such an optimization problem can be formulated as 
\begin{equation}
\underset{w}{\textup{min}}\frac{1}{N}\sum_{i=1}^{N}p_{i}L_{i}\left ( w_{i},D_{i}\right )
\end{equation}
where $L_i$is the local loss function at each client $C_i$ , $p_i$=$\frac{\left |D_i\right |}{\sum_{i=1}^{N} {\left| D_i \right |}}\geqslant 0$, with $\sum_{i=1}^{N}p_i$=$1$. Formally, to achieve global optimization, the server aggregates the weights received from $N$ clients as $w=\sum_{i=1}^{N}p_iw_i$ in $t$-th communication round. To solve the global optimization problem, we need $T$ communication rounds in total.

\subsection{Motivation}
The proposed method is based on an intuitive observation: the performance of a face recognition model is positively correlated to the number of identities (IDs) of training data, and thereby more discriminative feature representations can be generated by generic face recognition model with more training IDs. Despite the fact that the label information of the extensive IDs can improve local training with local data in the federated learning setup, the global shareable knowledge has not been fully leveraged by existing methods.
\par In this work, we expect to make use of the publicly available knowledge at the server during local training. The transmission of a huge amount of images from the server to clients meets memory and resource restrictions, and thus we propose to share mean features of these images from the same IDs for optimizing local updates, which significantly mitigates the communication and computation cost. Furthermore, with mean feature representations, local optimization aims to push apart feature embeddings from different classes. Meanwhile, to alleviate the excessive guidance of global shared mean features as negatives, we introduce a positive regularization to control the drift between the representations learned by the local model and the global model.

\subsection{Framework Overview}
\par Over the above insight, we present \textbf{AdaFedFR}, an efficient federated face recognition framework with the adaptive inter-class representation contrastive learning. Figure~\ref{fig2} illustrates an overview of our proposed framework. Instead of transmission of large-scale dataset, mean feature representations are shared from server to clients. For local training, we optimize the local model with a multi-objective, formalized by cosface loss, contrastive loss and bce loss. In addition, following \cite{wei2020federated}, differential privacy is added to the uploaded weights for a further privacy guarantee in real industrial applications. The details of AdaFedFR are summarized in the supplementary materials.
\subsection{Shareable Representations}
One of previous federated learning works in face recognition, FedFR\cite{liu2022fedfr}, has leveraged globally shared data with a hard negative sampling strategy in order to regularize the local training and prevent the model from overfitting to local identities. However, these techniques incur additional communication cost, and it is challenging to be applied on devices with limited memory size and computing power in real-world scenarios. To address this challenge, we propose a novelty transmission-friendly strategy by sharing normalized mean feature embeddings from all IDs (classes) at the server, which greatly degrades the transmission consumption compared with data sharing and plays an important role in regularizing the local updates. 
\par At the beginning of each communication round $t$, we first use global model $w_S^i$ to extract features of public data and calculate their mean feature representations matrix $\textbf{R}\in \mathbb{R}^{G\times d}$, where $G$ is the number of public identities at the server and $d$ is the dimension of features extracted from the model. The mean representations would be formulated as:
\begin{equation}
\textbf{R} = \frac{1}{\left |D_{S,i} \right |}\sum_{x\in D_{S,i}}^{}F\left ( x \right ), \end{equation}
where $F\left ( x \right )$ is the representations of public datasets $D_{S}$. The $i$ denotes the $i$-th class of data.

\subsection{Representation Learning in Feature Space}
The above strategy demonstrates the mean feature embeddings from public identities are considered as auxiliary knowledge to regularize the local training at clients. Intuitively, mean feature embedding represents the cluster center of the same class in feature space and samples from different classes should be pushed away from each other. Formally, we introduce adaptive inter-class representation learning to regularize the local updates with global class representations which are regarded as negatives in local contrastive objective.
\par Inspired by the N-pair loss objective\cite{sohn2016improved}, which proposed a scalable and effective deep metric learning objective that significantly improves contrastive learing by pushing away multiple negative examples jointly at each update. The N-pair loss term is defines as:
\begin{equation}
\mathcal{L}=-\log \frac{\exp \left(\operatorname{sim}\left(f, f_{pos}\right)\right)}{\exp \left(\operatorname{sim}\left(f, f_{pos}\right)\right)+\sum_{i}^{N} \exp \left(\operatorname{sim}\left(f, f_{neg, i}\right)\right)},\small 
\end{equation}
where $\operatorname{sim}(\cdot, \cdot)$ is the cosine similarity between features, and $N$ is the number of negative samples. We develop a novel representation learning approach in our federated face recognition task, which aims to utilize the similarity between representations to correct the local training at each client.
\par To degrade the similarity between representation of local data and many "negative" representations in embedding space, we first introduce global feature embeddings from different classes to the clients. Due to the presence of negative representations from public identities, the local model is further optimized to identify a positive sample from multiple negative samples, which can learn more powerful and discriminative representations. Moreover, the hard negative relationship is considered as an essential component to regularize the distance of representation learning. Therefore, we delicately design an adaptive hard sampling strategy of negative representations at each client. In the local training process, learnable parameters are adopted to the selection of negative pairs based on the cosine similarities between local anchors and negative embeddings. Besides, motivated by the well-performed face representations extracted by the pre-trained model with global larger-scale data, we learn from the recent work \cite{li2021model} to make use of model-contrastive learning to control the drift between the representations learned by the local and global models. Similar to N-pair loss function \cite{sohn2016improved}, we define our $k$-negative contrastive loss with learnable selection vectors as:
\begin{equation}
\mathcal{L}_{kcl}=-\log\frac{\exp\left(\operatorname{sim}\left(f, f_{g}\right)/\tau\right)}{\exp\left(\operatorname{sim}\left(f, f_{g}\right)/\tau\right)+\sum_{i}^{G}\phi_{i}\exp\left(\operatorname{sim}\left(f, f_{p, i}\right)/\tau\right)},\small 
\end{equation}
where $f_{g}$ is the representation extracted by the global model at client, and $f_{p, i}$ is the representation from the public sample at server which is regarded as a negative one. $\tau$ denotes a temperature hyper-parameter. $G$ means the number of public class representatives. $\phi$ is a learnable vector that its element values are between $(0,1)$ to select the local-relevant features. After the selection, we obtain the top $k$ similarities between the anchor and negative representations.

\subsection{Local Objective}

To further improve the recognition performance of local specific identities while maintaining the capabilities of the generic model, we propose a novel local adapter that is trained using binary cross-entropy (BCE) loss. Different from the FedFR\cite{liu2022fedfr}, which relies on a large number of negative sample images, our approach utilizes the feature embeddings to compensate for the limited amount of data available for training the local classifier. Furthermore, we introduce the local adapter to enhance the performance, which allows us to effectively leverage the feature embeddings and achieve improved results compared to FedFR\cite{liu2022fedfr}.

This adapter aims to distinguish positive representations obtained from the local extractor and the negatives received directly from public identities. Thus, We formalize our bce loss following\cite{wen2021sphereface2, liu2022fedfr}:

\begin{equation}\small 
\begin{split}
&\mathcal{L}_{bce}= \frac{\lambda }{s^{'} } \cdot \log(1+  \exp(-s^{'}\cdot g(cos(\xi_{i}))-m^{'})-b )\\+ 
 &\frac{1-\lambda }{s^{'} } \cdot \sum_{j\ne i} \log(1+  \exp(s^{'}\cdot g(cos(\xi_{j}))+m^{'})+b ),
\end{split}
\end{equation}
where $\xi_{i}, \xi_{j}$ represent the adapter output on the local extracted feature and the global representation, respectively. The notations $\lambda,s_{'},m_{'},b$ and $g$ all follow those in the related work, which are the balanced factor, scaling constant, cosine margin, learned bias and enhancement function.

\par Therefore, our method formulates a fused loss function of representation learning and face classifier learning paradigm. Formally, we define the overall loss function for the local objective as: 
\begin{equation}
\mathcal{L}_{\text {overall }}=\alpha_1 \mathcal{L}_{\text {lmc }}+\alpha_2 \mathcal{L}_{\text {kcl }}+\alpha_3 \mathcal{L}_{\text {bce }}, 
\end{equation} 
where $\alpha_1, \alpha_2$ and $\alpha_3$ denote the weights of objectives. In the local training, each client updates the local model based on local training data and global representations.
As shown in Figure~\ref{fig2}, our local loss function consists of three portions: the margin-based cosface loss, the adaptive $k$-negative contrastive loss and bce loss. The first one is a typical face recognition objective adapting margins in the softmax loss function, which adds the cosine margin penalties to enforce the inter-class diversity and intra-class compactness for maximizing class separability in local feature space. The second loss is proposed by us which can help local model learn a better feature space by introducing additional class representations $R$ from server and representations extracted by the global model $w^{t}$. With the representation learning, the local model can continuously improve the generic face recognition model and prevent the training from overfitting the objective of local limited classes.

\begin{table}
\caption{The comparison of 1:1 verification (TAR@FAR=1e-4 and TAR@FAR=1e-3) on IJB-B and IJB-C datasets with different methods.}
\centering
\scalebox{0.9}{
\begin{tabular}{p{2.8cm}<{\centering}p{1.8cm}<{\centering}p{1.8cm}<{\centering}p{1.8cm}<{\centering}p{1.8cm}<{\centering}}
\toprule[0.8pt]
\multirow{2}{*}{\textbf{Method}} & \multicolumn{2}{c}{\textbf{IJB-B}} & \multicolumn{2}{c}{\textbf{IJB-C}} \\\cline{2-5}
 & $1e$-$4$ & $1e$-$3$ & $1e$-$4$ & $1e$-$3$ \\\midrule[0.6pt]
FedAvg\cite{mcmahan2017communication} & 70.33 & 82.54 & 73.15 & 84.68\\
Moon\cite{li2021model}  & 69.01 & 81.91 & 74.02 & 85.42\\
FedFace\cite{aggarwal2021fedface} & 70.75 & 82.69& 74.39 & 85.46\\ 
FedGC\cite{niu2022federated} & 70.56 & 82.53 &75.01 & 85.69 \\
FedFR\cite{liu2022fedfr}  & 76.61 &  85.59 & 80.30 & 88.27\\\midrule[0.6pt]
\textbf{Ours} & \textbf{77.06} & \textbf{86.12} & \textbf{80.67} & \textbf{88.78}\\\bottomrule[0.8pt]
\end{tabular}
}
\label{tab1}
\vspace{-0.3cm}
\end{table}

\begin{table*} 
\caption{Ablation studies. The verification performance on the JIB-C dataset with different modules in federated learning settings. There are one server with 6k identities and 20 clients where each client contain 100 identities.}
\centering
\scalebox{0.85}{
\begin{tabular}{l|p{1.3cm}<{\centering}p{1.4cm}<{\centering}p{1.6cm}<{\centering}|p{1.4cm}<{\centering}p{1.0cm}<{\centering}|p{1.0cm}<{\centering}p{1.0cm}<{\centering}p{1.0cm}<{\centering}p{1.0cm}<{\centering}p{1.0cm}<{\centering}}
\toprule[1pt]
\multirow{3}{*}{\textbf{Setup}}& \multicolumn{3}{c|}{\textbf{Modules}} & \multicolumn{6}{c}{\textbf{IJB-C Evaluations}} \\\cline{2-10}
& \multirow{2}{*}{\makecell[c]{\small{$Glob Data$}}} & \multirow{2}{*}{\small{$Glob Feat$}} & \multirow{2}{*}{\makecell[c]{\small{$Contrastive$}}}  & \multicolumn{2}{c|}{$1:1$} & \multicolumn{4}{c}{$1:N$} \\\cline{5-10}
& & & & $1e$-$4$ & $1e$-$3$ & $1e$-$2$ & $1e$-$1$ & $Top1$ & $Top5$\\\midrule[0.6pt]
\multicolumn{4}{l|}{Centrally pre-trained on 6k IDs} & 78.42 & 87.67 & 60.32 & 70.82 & 83.44 & 89.50\\\cmidrule[0.6pt]{1-10}
\multirow{5}{*}{\makecell[l]{Federated\\Learning\\on 2k IDs}} & \tiny{\XSolidBrush}  & \tiny{\XSolidBrush}  & \tiny{\XSolidBrush}  & 73.15 & 84.68 & 52.23 & 65.09 & 83.22 & 89.08\\
& \tiny{\XSolidBrush} & \tiny{\XSolidBrush} & $\checkmark$  & 74.02 &  85.42 & 53.11 & 65.51 & 83.41 & 89.19\\ 
& $\checkmark$ & \tiny{\XSolidBrush} & \tiny{\XSolidBrush}  & 79.94 & 88.25 & 61.07 & 71.36 & 83.72 & 89.16\\ 
& $\checkmark$ & \tiny{\XSolidBrush} & $\checkmark$  & 79.96 & 88.14 & 62.69 & 72.50 & 83.83 & 89.68\\ 
& \tiny{\XSolidBrush} & $\checkmark$ & $\checkmark$ & \textbf{80.67} & \textbf{88.78} & \textbf{63.30} & \textbf{74.23} & \textbf{84.42} & \textbf{90.10} \\\midrule[0.6pt]
\multicolumn{4}{l|}{Centrally trained on 8k IDs}  & 81.40 & 89.09 & 67.26 & 77.02 & 86.40 & 91.14\\ 
\bottomrule[1pt]
\end{tabular}
}
\label{tab2}
\end{table*}

\vspace{-0.2cm}
\section{Experiments}
\subsection{Experimental Settings}
\noindent \textbf{Datasets} 
\indent Considering the limited training resource, we employ MS-Celeb-1M\cite{guo2016ms} as training set and a reliable subset containing 5.8{\em M} images from 86{\em K} ids of MS-Celeb-1M is collected. Following the federated learning setting in \cite{liu2022fedfr}, we manually select 8,000 identities from the dataset and each identity contains 100 face images. The training sets are divided into two parts, where 6,000 identities are used for pre-training the global model, and 2,000 identities are split into 20 parts equally at local clients. For testing, we evaluate the performance of our generic face model on the large-scale benchmark datasets, IJB-B\cite{whitelam2017iarpa} and IJB-C\cite{maze2018iarpa}. Also, we explore the following benchmark datasets, LFW\cite{huang2008labeled}, CFP-FP\cite{sengupta2016frontal}, and AgeDB\cite{moschoglou2017agedb}.

\noindent \textbf{Evaluation Metrics}  
\indent For a fair comparison of the generic face model, we use the same evaluation metrics as in FedFR\cite{liu2022fedfr}, which strictly follow the IJB-C evaluation protocol that is commonly used in the face recognition tasks. We report the true acceptance rates (TAR) at different false acceptance rates (FAR) for 1:1 verification protocol, and true positive identification rates (TPIR) at different false positive identification rates (FPIR) for 1:N identification protocol. In addition, we report the ablation studies with the top-1 and top-5 identification accuracy to show the performance of different modules, which is the correct proportion of the first-ranked samples retrieved based on the feature similarity.
\par \noindent \textbf{Implementation Details} 
\indent For data preprocessing, we use the same image preprocessing approach as Arcface\cite{deng2019arcface}, generating the face image cropped to size 112$\times$112 by using five facial landmarks predicted by face detector, and the image is normalized to [-1, 1]. For the face embedding network, we adopt ResNet34\cite{he2016deep} as the backbone. We set the momentum to 0.9 and weight decay to 5e$-$4. For the local training, we employ the stochastic gradient descent (SGD) optimizer and the same learning rate schedules for different federated learning approaches. For the federated learning setup, both the number of communication rounds and local epochs in each communication are set to 10. The hyper-parameters in our proposed method are empirically set, where temperature $\tau=0.5$, cosface $m=0.4$ and $s$=30. In addition, the $\alpha_1, \alpha_2$ and $\alpha_3$ of the overall loss function are empirically set to $1$, $5$ and $10$, respectively.

\subsection{Method Comparison}
\par We compare our method with a series of federated learning baselines, including conventional federated learning frameworks and recent works on the particular face recognition task. The representative approaches in the former category include the general method FedAvg\cite{mcmahan2017communication}, Moon\cite{li2021model}. The latter shows more customized frameworks on face recognition, including the earliest federated face recognition framework FedFace\cite{aggarwal2021fedface} and the latest approach FedFR\cite{liu2022fedfr}. We retrain the above methods as the same data selection as the AdaFedFR. Furthermore, we conduct extensive experiments to majorly compare our proposed method with FedFR for the reason that its framework focuses on applying federated learning in more realistic face recognition settings. 
\par \noindent \textbf{Results on IJB-B, IJB-C, LFW, CFP-FP and AgeDB} \noindent We test AdaFedFR and other methods on the test datasets, IJB-B, including 1,845 subjects with 21.8k still images which has 10,270 positive matches and 8M negative matches in the 1:1 verification, and IJB-C, containing 31,334 images of 3,531 identities as in \cite{maze2018iarpa} which has 19,557 positive matches and 15,638,932 negative matches in the 1:1 verification. On IJB-B and IJB-C datasets, we employ the MS1M dataset and the ResNet34 backbone for a fair comparison with other methods. As shown in Table~\ref{tab1}, AdaFedFR shows significant improvements compared to other approaches. Furthermore, AdaFedFR also achieves the highest accuracy on LFW\cite{huang2008labeled}, CFP-FP\cite{sengupta2016frontal}, and AgeDB\cite{moschoglou2017agedb}, reported in the supplementary material. 

\begin{table}
\vspace{-0.1cm}
\caption{The time per round of different approaches including both the local model training at clients and aggregation at the server.}
\vspace{-0.2cm}
\centering
\scalebox{0.9}{
\begin{tabular}{p{3cm}<{\centering}p{2.8cm}<{\centering}p{2.8cm}<{\centering}}
\toprule[1pt]
\textbf{Method} & \textbf{time/round} & \textbf{speedup} \\\midrule[0.6pt]
FedAvg\cite{mcmahan2017communication} & 37 min & 1$\times$ \\
Moon\cite{li2021model} & 41 min & 0.9$\times$ \\
FedFace\cite{aggarwal2021fedface} & 51 min & 0.75$\times$ \\
FedGC\cite{niu2022federated} & 47 min & 0.8$\times$ \\
FedFR\cite{liu2022fedfr} & 194 min & 0.2$\times$ \\\midrule[0.6pt]
\textbf{Ours}  & \textbf{39 min} & \textbf{0.95$\times$}  \\\bottomrule[1pt]
\end{tabular}
}
\label{tab3}
\vspace{-0.5cm}
\end{table}

\par \noindent \textbf{Comparison with FedFR}
\indent FedFR\cite{liu2022fedfr} is a recent federated learning framework focusing on the realistic face recognition applications with globally shared data. In each communication round, FedFR sends the global datasets (600,000 face images of 6k identities) and class embeddings of shared identities to each client and subsequently jointly utilizes the global and local data to train the local model. In addition, FedFR proposes the hard negative sampling strategy to reduce the computation cost of  the local training. In this section, we compare our proposed method with FedFR from three perspectives, performance, computation cost and communication efficiency. 

\begin{figure}[t]
\begin{center}
   \includegraphics[width=0.6\linewidth]{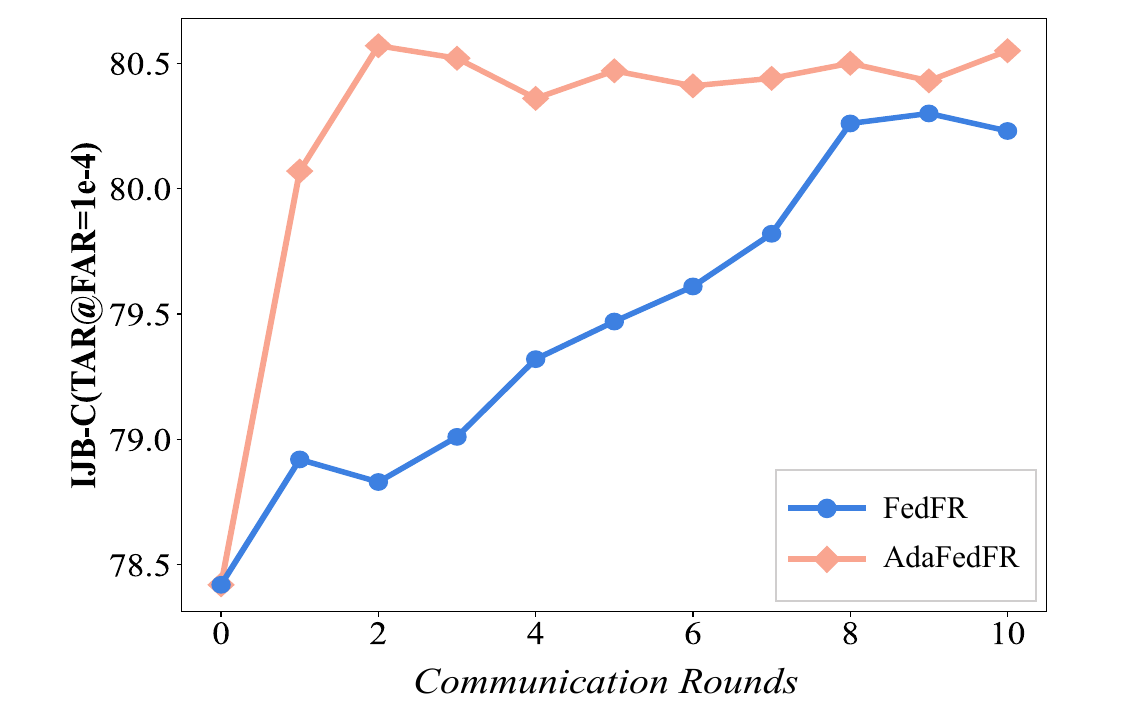}
\end{center}
\vspace{-0.4cm}
   \caption{The comparison of generic model performance under different communications round between FedFR\cite{liu2022fedfr} and our method. AdaFedFR achieves the outperforming performance in 2 communication rounds.}
\label{fig3}
\end{figure}

\begin{itemize}
\item \textbf{Performance.} To compare the experimental results with the FedFR, as shown in Table~\ref{tab1}, our proposed method boost the FedFR by $0.58\%$, $0.62\%$, $0.46\%$ and $0.58\%$ on both TAR@FAR=1e-4 and TAR@FAR=1e-3 for IJB-B and IJB-C, respectively. Besides the sharing global class representations, our method introduces the auxiliary relationship between negative representation pairs in feature space, which enables the generic model to learn more discriminative representations.

\item \textbf{Computation Cost.} We consider the FedAvg as the baseline, because both AdaFedFR and FedFR are the extension of the framework in FedAvg. The average training time per round experimented on 4 NVIDIA Tesla V100 GPUs with batch size 1024 is shown in Table~\ref{tab3}. Despite introducing the additional feature representation matrix and contrastive loss term in the local training phase, the training time of AdaFedFR would not be higher than FedAvg. In contrast, FedFR not only results in additional communication with sharing large-scale data, but also trains much slower. To sum up, our method speeds up the training of each communication round by $5\times$ faster than FedFR.
\item \textbf{Communication Efficiency.} The performance in each communication round during federated training is shown in Figure~\ref{fig3}. In this section, we conduct the training with the same 10 communication rounds for a fair comparison. The curves in Figure~\ref{fig3} show that the speed of improvement on performance in AdaFedFR is much faster than FedFR. Besides, our method could achieve outperforming performance in less than three communication rounds, which demonstrate the superiority of communication efficiency.
\end{itemize}
\label{sec:intro}
\vspace{-0.2cm}
Consequently, compared with FedFR, our proposed AdaFedFR can improve the generic face representation with much higher training efficiency.


\vspace{-0.4cm}
\subsection{Ablation Study}
\vspace{-0.3cm}
\indent We validate the effectiveness of different modules, including the global data, global class feature representations and contrastive learning progress. The ablation studies have been reported in Table~\ref{tab2}. We implement the extensive experiments with one central server and 20 clients, and there are 100 identities at each client. To compare the effectiveness of different modules, we conduct the validation for both the 1:1 verification and 1:N identification on the IJB-C dataset.
\par In Table~\ref{tab2}, the first row is the performance of pre-training model on centralized 6k identities. The pre-training model is the baseline for each modules to optimize in federated learning setup. Moreover, the last row is the performance of centrally training model on total 8k identities, which can be regarded as the up-board of federated learning optimization. The main ablation experiments are shown from the second to the sixth row in Table~\ref{tab2}. In the second row, we utilize the baseline approach FedAvg to optimize the generic model, and the performance is worse than the pre-training model due to the overfitting to the local data. 



Then, we introduce the model-contrastive federated learning to further correct the local updates by maximizing the agreement of representation learned by the local model and the representation learned by the global model. To begin with, we only conduct the model-contrastive learning on our experiment settings, which provides little improvement over FedAvg. The result shows that the compelling work in image classification task cannot be directly applied onto face recognition. 
To demonstrate the effectiveness of combination of additional global knowledge and contrastive objective, we attempt to leverage the public shared data available from web sites. Follow the data sharing setting as FedFR, we use the pretrained dataset as the global data in our experiments.

\par In addition, we further evaluate the effectiveness of our proposed module, consisting of global shared feature representations of identities and proposed adaptive $k$-negative contrastive learning algorithm. In the sixth row, we can see the overall performance of generic model, which achieves superior results of all evaluations on IJB-C. Moreover, the other ablation studies, (e.g., ablations of hard negatives \ref{subsec:hard} , privacy analysis \ref{subsec:pri}, and personalized evaluation \ref{subsec:per}), are showed in the supplemental material.

\vspace{-0.4cm}
\section{Conclusion}
\vspace{-0.4cm}
This paper has proposed a simple yet elegant federated face recognition framework based on adaptive inter-class representation learning, called AdaFedFR. We address the face recognition model training under the practical federated learning setting and start from a novel perspective of sharing identity feature representations from server to each client, fruitfully adopting representation learning to optimize generic face model and largely alleviates the additional computational consumption in comparison with previous works. Besides, AdaFedFR is a communication-friendly framework for realistic face recognition applications under differential privacy guarantee, achieving high recognition accuracy in less than three communication rounds. Extensive experimental results on public benchmarks further demonstrate that our method has excellent performance in terms of both accuracy and efficiency.


\begin{thebibliography}{10}

\bibitem{aggarwal2021fedface}
Divyansh Aggarwal, Jiayu Zhou, and Anil~K Jain.
\newblock Fedface: Collaborative learning of face recognition model.
\newblock In {\em 2021 IEEE International Joint Conference on Biometrics (IJCB)}, pages 1--8. IEEE, 2021.

\bibitem{an2021partial}
Xiang An, Xuhan Zhu, Yuan Gao, Yang Xiao, Yongle Zhao, Ziyong Feng, Lan Wu, Bin Qin, Ming Zhang, Debing Zhang, et~al.
\newblock Partial fc: Training 10 million identities on a single machine.
\newblock In {\em Proceedings of the IEEE/CVF International Conference on Computer Vision}, pages 1445--1449, 2021.

\bibitem{cao2018vggface2}
Qiong Cao, Li~Shen, Weidi Xie, Omkar~M Parkhi, and Andrew Zisserman.
\newblock Vggface2: A dataset for recognising faces across pose and age.
\newblock In {\em 2018 13th IEEE international conference on automatic face \& gesture recognition (FG 2018)}, pages 67--74. IEEE, 2018.

\bibitem{deng2020sub}
Jiankang Deng, Jia Guo, Tongliang Liu, Mingming Gong, and Stefanos Zafeiriou.
\newblock Sub-center arcface: Boosting face recognition by large-scale noisy web faces.
\newblock In {\em European Conference on Computer Vision}, pages 741--757. Springer, 2020.

\bibitem{deng2019arcface}
Jiankang Deng, Jia Guo, Niannan Xue, and Stefanos Zafeiriou.
\newblock Arcface: Additive angular margin loss for deep face recognition.
\newblock In {\em Proceedings of the IEEE/CVF conference on computer vision and pattern recognition}, pages 4690--4699, 2019.

\bibitem{deng2021variational}
Jiankang Deng, Jia Guo, Jing Yang, Alexandros Lattas, and Stefanos Zafeiriou.
\newblock Variational prototype learning for deep face recognition.
\newblock In {\em Proceedings of the IEEE/CVF Conference on Computer Vision and Pattern Recognition}, pages 11906--11915, 2021.

\bibitem{deng2017marginal}
Jiankang Deng, Yuxiang Zhou, and Stefanos Zafeiriou.
\newblock Marginal loss for deep face recognition.
\newblock In {\em Proceedings of the IEEE conference on computer vision and pattern recognition workshops}, pages 60--68, 2017.

\bibitem{guo2016ms}
Yandong Guo, Lei Zhang, Yuxiao Hu, Xiaodong He, and Jianfeng Gao.
\newblock Ms-celeb-1m: A dataset and benchmark for large-scale face recognition.
\newblock In {\em European conference on computer vision}. Springer, 2016.

\bibitem{he2016deep}
Kaiming He, Xiangyu Zhang, Shaoqing Ren, and Jian Sun.
\newblock Deep residual learning for image recognition.
\newblock In {\em Proceedings of the IEEE conference on computer vision and pattern recognition}, pages 770--778, 2016.

\bibitem{huang2008labeled}
Gary~B Huang, Marwan Mattar, Tamara Berg, and Eric Learned-Miller.
\newblock Labeled faces in the wild: A database forstudying face recognition in unconstrained environments.
\newblock In {\em Workshop on faces in'Real-Life'Images: detection, alignment, and recognition}, 2008.

\bibitem{huang2020curricularface}
Yuge Huang, Yuhan Wang, Ying Tai, Xiaoming Liu, Pengcheng Shen, Shaoxin Li, Jilin Li, and Feiyue Huang.
\newblock Curricularface: adaptive curriculum learning loss for deep face recognition.
\newblock In {\em proceedings of the IEEE/CVF conference on computer vision and pattern recognition}, pages 5901--5910, 2020.

\bibitem{karimireddy2020scaffold}
Sai~Praneeth Karimireddy, Satyen Kale, Mehryar Mohri, Sashank Reddi, Sebastian Stich, and Ananda~Theertha Suresh.
\newblock Scaffold: Stochastic controlled averaging for federated learning.
\newblock In {\em International Conference on Machine Learning}, pages 5132--5143. PMLR, 2020.

\bibitem{li2021model}
Qinbin Li, Bingsheng He, and Dawn Song.
\newblock Model-contrastive federated learning.
\newblock In {\em Proceedings of the IEEE/CVF Conference on Computer Vision and Pattern Recognition}, pages 10713--10722, 2021.

\bibitem{li2021survey}
Qinbin Li, Zeyi Wen, Zhaomin Wu, Sixu Hu, Naibo Wang, Yuan Li, Xu~Liu, and Bingsheng He.
\newblock A survey on federated learning systems: vision, hype and reality for data privacy and protection.
\newblock {\em IEEE Transactions on Knowledge and Data Engineering}, 2021.

\bibitem{li2020federated}
Tian Li, Anit~Kumar Sahu, Manzil Zaheer, Maziar Sanjabi, Ameet Talwalkar, and Virginia Smith.
\newblock Federated optimization in heterogeneous networks.
\newblock {\em Proceedings of Machine Learning and Systems}, 2:429--450, 2020.

\bibitem{liu2022fedfr}
Chih-Ting Liu, Chien-Yi Wang, Shao-Yi Chien, and Shang-Hong Lai.
\newblock Fedfr: Joint optimization federated framework for generic and personalized face recognition.
\newblock In {\em Proceedings of the AAAI Conference on Artificial Intelligence}, volume~36, pages 1656--1664, 2022.

\bibitem{liu2017sphereface}
Weiyang Liu, Yandong Wen, Zhiding Yu, Ming Li, Bhiksha Raj, and Le~Song.
\newblock Sphereface: Deep hypersphere embedding for face recognition.
\newblock In {\em Proceedings of the IEEE conference on computer vision and pattern recognition}, pages 212--220, 2017.

\bibitem{liu2016large}
Weiyang Liu, Yandong Wen, Zhiding Yu, and Meng Yang.
\newblock Large-margin softmax loss for convolutional neural networks.
\newblock {\em arXiv preprint arXiv:1612.02295}, 2016.

\bibitem{masi2018deep}
Iacopo Masi, Yue Wu, Tal Hassner, and Prem Natarajan.
\newblock Deep face recognition: A survey.
\newblock In {\em 2018 31st SIBGRAPI conference on graphics, patterns and images (SIBGRAPI)}, pages 471--478. IEEE, 2018.

\bibitem{maze2018iarpa}
Brianna Maze, Jocelyn Adams, James~A Duncan, Nathan Kalka, Tim Miller, Charles Otto, Anil~K Jain, W~Tyler Niggel, Janet Anderson, Jordan Cheney, et~al.
\newblock Iarpa janus benchmark-c: Face dataset and protocol.
\newblock In {\em 2018 international conference on biometrics (ICB)}, pages 158--165. IEEE, 2018.

\bibitem{mcmahan2017communication}
Brendan McMahan, Eider Moore, Daniel Ramage, Seth Hampson, and Blaise~Aguera y~Arcas.
\newblock Communication-efficient learning of deep networks from decentralized data.
\newblock In {\em Artificial intelligence and statistics}, pages 1273--1282. PMLR, 2017.

\bibitem{meng2022improving}
Qiang Meng, Feng Zhou, Hainan Ren, Tianshu Feng, Guochao Liu, and Yuanqing Lin.
\newblock Improving federated learning face recognition via privacy-agnostic clusters.
\newblock {\em arXiv preprint arXiv:2201.12467}, 2022.

\bibitem{moschoglou2017agedb}
Stylianos Moschoglou, Athanasios Papaioannou, Christos Sagonas, Jiankang Deng, Irene Kotsia, and Stefanos Zafeiriou.
\newblock Agedb: the first manually collected, in-the-wild age database.
\newblock In {\em proceedings of the IEEE conference on computer vision and pattern recognition workshops}, pages 51--59, 2017.

\bibitem{nech2017level}
Aaron Nech and Ira Kemelmacher-Shlizerman.
\newblock Level playing field for million scale face recognition.
\newblock In {\em Proceedings of the IEEE Conference on Computer Vision and Pattern Recognition}, pages 7044--7053, 2017.

\bibitem{niu2022federated}
Yifan Niu and Weihong Deng.
\newblock Federated learning for face recognition with gradient correction.
\newblock In {\em Proceedings of the AAAI Conference on Artificial Intelligence}, volume~36, pages 1999--2007, 2022.

\bibitem{schroff2015facenet}
Florian Schroff, Dmitry Kalenichenko, and James Philbin.
\newblock Facenet: A unified embedding for face recognition and clustering.
\newblock In {\em Proceedings of the IEEE conference on computer vision and pattern recognition}, pages 815--823, 2015.

\bibitem{sengupta2016frontal}
Soumyadip Sengupta, Jun-Cheng Chen, Carlos Castillo, Vishal~M Patel, Rama Chellappa, and David~W Jacobs.
\newblock Frontal to profile face verification in the wild.
\newblock In {\em 2016 IEEE winter conference on applications of computer vision (WACV)}, pages 1--9. IEEE, 2016.

\bibitem{sohn2016improved}
Kihyuk Sohn.
\newblock Improved deep metric learning with multi-class n-pair loss objective.
\newblock {\em Advances in neural information processing systems}, 29, 2016.

\bibitem{sun2014deep}
Yi~Sun, Yuheng Chen, Xiaogang Wang, and Xiaoou Tang.
\newblock Deep learning face representation by joint identification-verification.
\newblock {\em Advances in neural information processing systems}, 27, 2014.

\bibitem{taigman2014deepface}
Yaniv Taigman, Ming Yang, Marc'Aurelio Ranzato, and Lior Wolf.
\newblock Deepface: Closing the gap to human-level performance in face verification.
\newblock In {\em Proceedings of the IEEE conference on computer vision and pattern recognition}, pages 1701--1708, 2014.

\bibitem{terhorst2020ser}
Philipp Terhorst, Jan~Niklas Kolf, Naser Damer, Florian Kirchbuchner, and Arjan Kuijper.
\newblock Ser-fiq: Unsupervised estimation of face image quality based on stochastic embedding robustness.
\newblock In {\em Proceedings of the IEEE/CVF conference on computer vision and pattern recognition}, pages 5651--5660, 2020.

\bibitem{voigt2017eu}
Paul Voigt and Axel Von~dem Bussche.
\newblock The eu general data protection regulation (gdpr).
\newblock {\em A Practical Guide, 1st Ed., Cham: Springer International Publishing}, 10(3152676):10--5555, 2017.

\bibitem{wang2018additive}
Feng Wang, Jian Cheng, Weiyang Liu, and Haijun Liu.
\newblock Additive margin softmax for face verification.
\newblock {\em IEEE Signal Processing Letters}, 25(7):926--930, 2018.

\bibitem{wang2018cosface}
Hao Wang, Yitong Wang, Zheng Zhou, Xing Ji, Dihong Gong, Jingchao Zhou, Zhifeng Li, and Wei Liu.
\newblock Cosface: Large margin cosine loss for deep face recognition.
\newblock In {\em Proceedings of the IEEE conference on computer vision and pattern recognition}, pages 5265--5274, 2018.

\bibitem{wei2020federated}
Kang Wei, Jun Li, Ming Ding, Chuan Ma, Howard~H Yang, Farhad Farokhi, Shi Jin, Tony~QS Quek, and H~Vincent Poor.
\newblock Federated learning with differential privacy: Algorithms and performance analysis.
\newblock {\em IEEE Transactions on Information Forensics and Security}, 15:3454--3469, 2020.

\bibitem{wen2021sphereface2}
Yandong Wen, Weiyang Liu, Adrian Weller, Bhiksha Raj, and Rita Singh.
\newblock Sphereface2: Binary classification is all you need for deep face recognition.
\newblock {\em arXiv preprint arXiv:2108.01513}, 2021.

\bibitem{whitelam2017iarpa}
Cameron Whitelam, Emma Taborsky, Austin Blanton, Brianna Maze, Jocelyn Adams, Tim Miller, Nathan Kalka, Anil~K Jain, James~A Duncan, Kristen Allen, et~al.
\newblock Iarpa janus benchmark-b face dataset.
\newblock In {\em proceedings of the IEEE conference on computer vision and pattern recognition workshops}, pages 90--98, 2017.

\bibitem{yu2020federated}
Felix Yu, Ankit~Singh Rawat, Aditya Menon, and Sanjiv Kumar.
\newblock Federated learning with only positive labels.
\newblock In {\em International Conference on Machine Learning}, pages 10946--10956. PMLR, 2020.

\bibitem{zhu2021webface260m}
Zheng Zhu, Guan Huang, Jiankang Deng, Yun Ye, Junjie Huang, Xinze Chen, Jiagang Zhu, Tian Yang, Jiwen Lu, Dalong Du, et~al.
\newblock Webface260m: A benchmark unveiling the power of million-scale deep face recognition.
\newblock In {\em Proceedings of the IEEE/CVF Conference on Computer Vision and Pattern Recognition}, pages 10492--10502, 2021.

\end{thebibliography}

\newpage
\section{Supplemental Material}

\subsection{Framework Details}\label{sec:train-recipe}

The specifics of AdaFedFR are presented to afford an accurate comprehension of the framework, as shown in Algorithm \ref{alg}.

\begin{algorithm}[t]
 \caption{The AdaFedFR framework}
 \begin{algorithmic}[1]
 \renewcommand{\algorithmicrequire}{\textbf{Input:}}
 \renewcommand{\algorithmicensure}{\textbf{Output:}}
    
    \Require Public datasets $\mathcal{D}_S$, local datasets $\mathcal{D}_i$, number of communication rounds $T$, number of local epochs $E$, number of global classes $G$, number of clients $N$, learning rate $\eta $.
    \Ensure The final model parameters $w$

 \State \textbf{\underline{Server executes:}}
 \State initialize $w^0$, $z^0$
    \For{$t = 0, 1, ..., T-1$}
        \State extract feature representations $\textbf{R}^{t}$ with $w^t$:
            \begin{center}
                $\textbf{R}^{t}= \bigcup_{i\in G}^{}r_{i}= \frac{1}{\left |D_{S,i} \right |}\sum_{x\in D_{S,i}}^{}F\left ( x \right )$
            \end{center} 
		\For{$i = 1, 2, ..., N$ \textbf{in parallel}}
			\State send $w^t$ and $\textbf{R}^{t}$ to $C_i$
			\State $\tilde{w}_{i}^{t+1} \leftarrow$ \textbf{ClientLocalUpdates}($i$, $t$, $w^t$, $\textbf{R}^{t}$)

		\EndFor
	
		\State $w_{i}^{t+1} \leftarrow \sum_{i=1}^N \frac{\left | \mathcal{D}_i \right |}{\left | \mathcal{D} \right |}\tilde{w}_{i}^{t+1}$
	\EndFor
	\State return $w$

\vspace*{0.3\baselineskip} 
\State \textbf{\underline{ClientLocalUpdates}}($i$, $t$, $w^t$, $\textbf{R}^{t}$)\textbf{:}
	\State $(w_{i}^{t}, z_{i}^{t}, \theta_{i}^{t}, \phi_{i}^{t}) \leftarrow w^t, z_{i}^{t-1}, \theta_{i}^{t-1}, \phi_{i}^{t-1}$
	\For{$epoch = 1,2, ..., E$}
	
	    \State Calculate total loss $\mathcal{L}_{\text {overall }}$
		\State Update the local extractor and the embedding
		    \State$z_{i}^{t}\leftarrow z_{i}^{t}-\eta \bigtriangledown _{z_{i}^{t}  } \mathcal{L}_{lmc}$
                \State$\theta_{i}^{t}\leftarrow \theta_{i}^{t}-\eta \bigtriangledown _{\theta_{i}^{t}  } \mathcal{L}_{bce}$
                 \State$w_{i}^{t}\leftarrow w_{i}^{t}-\eta \bigtriangledown _{w_{i}^{t}  } (\mathcal{L}_{lmc}+\mathcal{L}_{kcl})$
                \State$\phi_{i}^{t}\leftarrow \phi_{i}^{t}-\eta \bigtriangledown _{\phi_{i}^{t}  } \mathcal{L}_{overall}$
		\State Clip the local parameters 
		\par \State $w_{i}^{t+1}=w_{i}^{t+1}/\textup{max}\left ( a,\frac{\left \| w_{i}^{t+1} \right \|} {\mathcal H} \right )$
	\EndFor
	\State Add noise to uploading parameters $\tilde{w}_{i}^{t+1}=w_{i}^{t+1}+n_i$
	\State return $\tilde{w}_{i}^{t+1}$ to the server

 \end{algorithmic}
 \label{alg}
\end{algorithm}

\subsection{Ablations of Hard Negatives}\label{subsec:hard}
In our federated learning framework, the feature representations of different identities on the server are shared to each client, considered as negative representations to correct the local update in federated learning settings. Firstly, we calculate the pair-wise cosine similarities between the normalized local representation learned by local model and global negative class representations. And we sample the hard negative representations based on cosine similarities that the higher means the harder to be learned. The results are shown in Table~\ref{tab5}, the performance of AdaFedFR cannot be increased by increasing the number of negative pairs. In addition, we believe the best choice of negative pairs number on different datasets depends on the inherent feature space of the pre-trained model. Therefore, we devise a learnable k and achieve the best performance which is also presented in our main paper.  

\vspace{-0.3cm}
\begin{table}[h]
\centering
\caption{The influence of number of negative representations on IJB-B and IJB-C datasets.}
\vspace{-0.1cm}
\begin{tabular}{p{3cm}<{\centering}p{1.5cm}<{\centering}p{1.5cm}<{\centering}p{1.5cm}<{\centering}p{1.5cm}<{\centering}}
\toprule[1pt]
\multirow{1}{*}{\textbf{\makecell[c]{Number \\  of Negatives}}} & \multicolumn{2}{c}{\textbf{IJB-B}} & \multicolumn{2}{c}{\textbf{IJB-C}} \\\cline{2-5}
 & $1e$-$4$ & $1e$-$3$ & $1e$-$4$ & $1e$-$3$ \\\midrule[0.6pt]
3 & 75.29 & 85.37 & 79.43 & 88.01 \\
5 & 76.71 & 85.88 & 80.55 & 88.66\\ 
10 & 76.89 & 85.67 & 80.63 & 88.45\\ 
20 & 76.86 & 85.73 &79.88 & 88.33\\
100 & 74.06 & 84.74& 78.50 & 87.77\\ 
\textbf{learnable $k$} & \textbf{77.06} & \textbf{86.12} & \textbf{80.67} & \textbf{88.78}\\ 
\bottomrule[1pt]
\end{tabular}
\label{tab5}
\end{table}

\subsection{Privacy Analysis}\label{subsec:pri}
\indent Our proposed framework added proper gaussian perturbations to the transmitting parameters to satisfy the requirement of $\left ( \epsilon ,\delta  \right ) $-differential privacy. \cite{meng2022improving} demonstrated that with a specified $\left (\epsilon ,\delta\right ) $,  the larger number of communication rounds $T$ can lead to a higher chance of information leakage. Therefore, the faster convergence (i.e. smaller $T$) makes our framework achieve higher privacy preservation compared with previous works. Besides, it is worth noting that global class representations in AdaFedFR are extracted from publicly available datasets without privacy issues, which would not lead to privacy leakage. Additionally, considering that federated face recognition framework would meet tighter privacy restrictions in the future, gaussian noise can be added to the sharing class centers in AdaFedFR to satisfy differential privacy of individual record, which can further strengthen the privacy protection.
\vspace{-0.2cm}

\subsection{Personalized Evaluation}\label{subsec:per}
To validate the personalized performance of our proposed framework and local adapter(LA), we compare our method with FedFR\cite{liu2022fedfr} and the fine-tune method of the local model at clients. Following the experimental setup of FedFR, for each identity in each local client, we use 60 images for local training, and 40 images for personalized model evaluation, respectively. The FedFR optimizes personalized models for the corresponding clients via the proposed decoupled feature customization module. For fine-tune method, we firstly carry out the collaborative training between server and clients, to obtain a better generic model. Secondly, each client optimizes its local personalized model on basis of the pretrained generic model at the first stage with the local data, separately. As shown in Table~\ref{tab4}, our proposed framework with local adapter can outperform the FedFR and the local adaptation method.

\begin{table}[h]
\vspace{-0.2cm}
\caption{The comparison of personalized method.}
\vspace{-0.2cm}
\centering
\scalebox{0.95}{
\begin{tabular}{p{3cm}<{\centering}p{1.8cm}<{\centering}p{1.8cm}<{\centering}p{1.8cm}<{\centering}p{1.8cm}<{\centering}}
\toprule[0.8pt]
\multirow{3}{*}{\textbf{\makecell[c]{Method}}} & \multicolumn{4}{c}{\textbf{Personalized Evaluation}} \\\cline{2-5}
 & \multicolumn{2}{c}{1:1 TAR @ FAR} & \multicolumn{2}{c}{1:N TPIR @ FPIR} \\\cline{2-5}
  & $1e$-$6$ & $1e$-$5$ & $1e$-$5$ & $1e$-$4$ \\\midrule[0.6pt]
Fine-tune & 73.77 & 85.70 & 88.17 & 93.72\\
FedFR\cite{liu2022fedfr} & 88.24 & 95.78 & 95.12 & 98.45 \\\midrule[0.6pt]
\textbf{AdaFedFR} & \textbf{89.06} & \textbf{96.12} & \textbf{95.33} & \textbf{98.94}\\ \bottomrule[0.8pt]
\end{tabular}
}
\label{tab4}
\vspace{-0.5cm}
\end{table}

\subsection{Results on LFW, CFP-FP and AgeDB}\label{subsec:res}
\vspace{-0.5cm}
\begin{table}[H]
\centering
\caption{Comparison of the face recognition accuracy among various benchmarks.}
\begin{tabular}{p{3cm}<{\centering}p{2cm}<{\centering}p{2cm}<{\centering}p{2cm}<{\centering}}
\toprule[1pt]
\textbf{Method} & \textbf{LFW} & \textbf{CFP-FP}  & \textbf{AgeDB} \\\midrule[0.6pt]
FedAvg\cite{mcmahan2017communication} & 95.08 & 78.71 & 87.93\\
Moon\cite{li2021model} & 97.13 & 79.09 & 88.81\\
FedFace\cite{aggarwal2021fedface} & 97.34 & 79.31 & 88.68\\
FedGC\cite{niu2022federated} & 97.27 & 80.10 & 88.72 \\
FedFR\cite{liu2022fedfr} & 98.90 & 81.26 & 90.33 \\\midrule[0.6pt]
\textbf{Ours} &\textbf{98.97} & \textbf{83.34} & \textbf{90.56} \\\bottomrule[1pt]
\end{tabular}
\label{tab6}

\end{table}

 We also explore the following benchmark datasets, LFW\cite{huang2008labeled}, CFP-FP\cite{sengupta2016frontal}, and AgeDB\cite{moschoglou2017agedb}. As shown in Table~\ref{tab6}, the results demonstrate that our method outperforms previous approaches.


\end{document}